# Explainable Signature-based Machine Learning Approach for Identification of Faults in Grid-Connected Photovoltaic Systems

Syed Wali
*Department of Electrical and Computer Engineering*
*Texas A&M University*
College Station, TX, USA
syedwali@tamu.edu

Irfan Khan
*Marine Engineering Technology Department in a joint appointment with Electrical and Computer Engineering*
*Texas A&M University*
College Station, TX, USA
irfankhan@tamu.edu

***Abstract*—Transformation of conventional power networks into smart grids with the heavy penetration level of renewable energy resources, particularly grid-connected Photovoltaic (PV) systems, has increased the need for efficient fault identification systems. Malfunctioning any single component in grid-connected PV systems may lead to grid instability and other serious consequences, showing that a reliable fault identification system is the utmost requirement for ensuring operational integrity. Therefore, this paper presents a novel fault identification approach based on statistical signatures of PV operational states. These signatures are unique because each fault has a different nature and distinctive impact on the electrical system. Thus, the Random Forest Classifier trained on these extracted signatures showed 100% accuracy in identifying all types of faults. Furthermore, the performance comparison of the proposed framework with other Machine Learning classifiers depicts its credibility. Moreover, to elevate user trust in the predicted outcomes, SHAP (Shapley Additive Explanation) was utilized during the training phase to extract a complete model response (global explanation). This extracted global explanation can help in the assessment of predicted outcomes' credibility by decoding each prediction in terms of features contribution. Hence, the proposed explainable signature-based fault identification technique is highly credible and fulfills all the requirements of smart grids.**

## I. INTRODUCTION

Exponential increment in the power demand with the population growth and environmental changes highlights the need to integrate renewable energy resources in the conventional power network. There are several renewable resources that can be utilized for meeting the increased power demand. Among these resources, Photovoltaic Systems (PV) have the highest capacity of power generation and can be easily integrated with the power grid. Research and development in the Power electronics domain have made this integration, more convenient and reliable by introducing energy-efficient power converters [1]. These converters act as an interface between the power grid and PV system by transforming the DC power into AC while maintaining the power requirements of the grid [2].

The Grid-connected PV systems operate in an extremely vulnerable environment and are subjected to thermal and electrical stresses, arising from abnormal behavior of any electrical component involved in this integration. These problems in the PV system may lead to downtime and decrement in its operational reliability. For enhancing the operational integrity and reliability, analysis of the PV systems' operational parameters is crucial because this analysis can prevent the PV system from undesirable operational states [3].

Traditionally, site engineers used manual assessment approaches for planning the preventive and corrective maintenance of PV systems. However, the introduction of Machine Learning (ML) techniques has transformed this manual assessment process into more reliable programmable approaches. These approaches have shown outclass performance in several classification and pattern recognition problems [4] and can be utilized in analyzing problems and faults of the grid-connected PV systems.

The authors of [1] utilized a computationally expensive 25 layered deep neural architecture which is commonly known as Alex Net. They convert their acquired one-dimensional time-series signal into a two-dimensional image by passing through different preprocessing stages. First, they normalized their data and then divide it into equal segments such that these segments overlap each other. Then these fixed-length segments are transformed into the frequency domain and this two-dimensional spectrogram is used for training and testing of the Alex Net model. Furthermore, four different operational states of PV systems were tested in their research work and their model showed 98.3% accuracy on the test set.

Similarly, the Authors of [5] introduced a 2-dimensional Convolution Neural Network approach for detecting the faults in the PV system. This Deep Neural architecture showed an accuracy of 73.53% in detecting the faults of the test dataset. Contrary to the two-dimensional approach, the authors of [6] utilized a fundamental Artificial Neural Network (ANN) for the identification of partial shading problems in PV systems. Similarly, authors of [7, 8] proposed their fault identification system based on ANN. However, these researchers did not focus on the transparency of the decision-making approach which is inherently impossible for deep neural networks as they are referred as black-box models [9]. Moreover, Deep architectures require a large amount of data for training the weights of the model and the selection of numerous hyperparameters for DNN is also a challenging task.

This paper presents a novel statistical signature-based fault identification approach using an ensemble classifier. Random Forest Classifiers (RFC) are relatively simpler than DNN and are intrinsically interpretable. This capability of interpretation is utilized to enhance the user trust by providing explanations of each predicted outcome using SHAP (Shapley Additive Explanation) [10]. Since extraction of statistical parameters and training of pruned RFC model is relatively simple and computationally inexpensive, the proposed fault identification system can be easily integrated into grid-connected PV systems. Furthermore, the concept of decoding all predicted outcomes can assist the power operators in initiating the effective downstream measures for stabilizing the power network during any contingency and can also elevate the credibility of the classifier.

A brief introduction of the proposed fault identification system along with the related work has been presented in this section. The rest of the paper is organized as follows. Section II provides the details of the proposed fault identification system and steps involved in its development. Section III evaluates the performance of the proposed framework on the test dataset and compared it with other renowned ML-based classifiers. Finally, Section IV concludes the research paper.

## II. Proposed Framework

Deep neural network-based architectures not only require a large amount of training data but also lack transparency. Therefore, deployment of these architectures with PV arrays is difficult as they require more computational power resulting in a high initial cost. Considering these bottlenecks of the fault identification system, this paper presents an efficient and reliable signature-based interpretable approach for the identification of persisting problems in PV systems. This approach is illustrated in Fig. 1, depicting that it involves two different stages. The first one is the development phase of the fault identification system which is explained in this section, and the second one is the evaluation phase that assesses the performance of the proposed system.

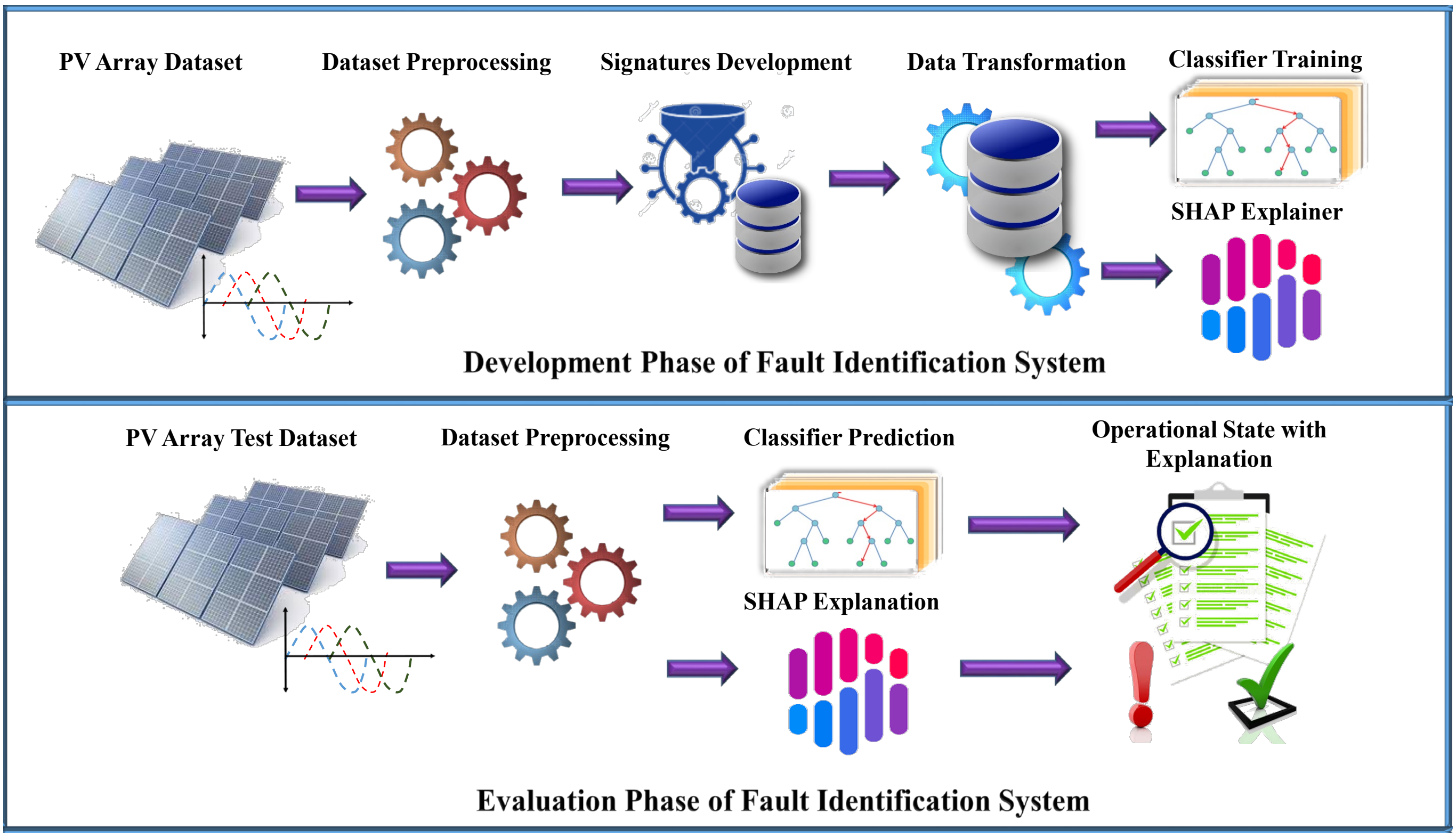

Fig. 1. Development and Evaluation Phase of Fault identification framework for grid connect PV systems. The development phase transforms the original GPVS dataset into a statistical signature-based data frame. These operational signatures of grid-connected PV systems are utilized in training the Random Forest Classifier and for extracting model global explanations using the SHAP framework. This trained model is evaluated on a test dataset for predicting faults in the system and explanations of each predicted outcome are obtained through SHAP during the evaluation phase.

The Grid-connected PV System Faults (GPVS-Faults) data by [11] is acquired by simulating different fault conditions in a lab-based PV microgrid system. This dataset contains the following seven different types of faults for Maximum Power Point Tracking mode (MPPT) and Intermediate Power Point Tracking (IPPT) mode of the PV system:

- Inverter Fault (F1): The first type of fault in the GPVS dataset is known as inverter fault which is caused by the failure of IGBTs. This type of fault may result in severe consequences and its earliest detection is the utmost requirement of PV systems.
- Feedback Sensor Fault (F2): The second type of fault in the GPVS dataset is based on malfunctioning of feedback sensor, reporting one of the phase voltages in a PV system.
- Grid Anomaly (F3): The third type of fault in the GPVS dataset is based on intermittent voltage fluctuations on the grid side that may bring operational instability.
- PV Array Mismatch (F4): Fourth type of fault in the GPVS dataset is based on a partial shading problem that reduces the performance of the PV system. This fault is less severe and difficult to detect [11].
- PV Array Mismatch (F5): The fifth type of fault in the GPVS dataset is based on a 15% open circuit problem of PV array and is difficult to detect due to its incipient nature.

- Controller Fault (F6): Sixth type of fault in the GPVS dataset is based on the problem in a gain of the PI controller of the boost converter.
- Boost Converter fault (F7): The seventh type of fault in the GPVS dataset is based on the problem in the time constant of the PI controller of the boost converter.

The sampling frequency of acquired signals in this dataset is 100 µseconds, which means that the consecutive 200 samples reflect one complete waveform. Furthermore, thirteen fault-related variables (Vpv, Ipv, Vdc, ia, ib, ic, va, vb, vc, |Iabc|, |Vabc|, fI, fV), as shown in Fig. 2 is provided in this dataset for the development of PV fault detection system and quantitative description of selected dataset files is provided in Table I.

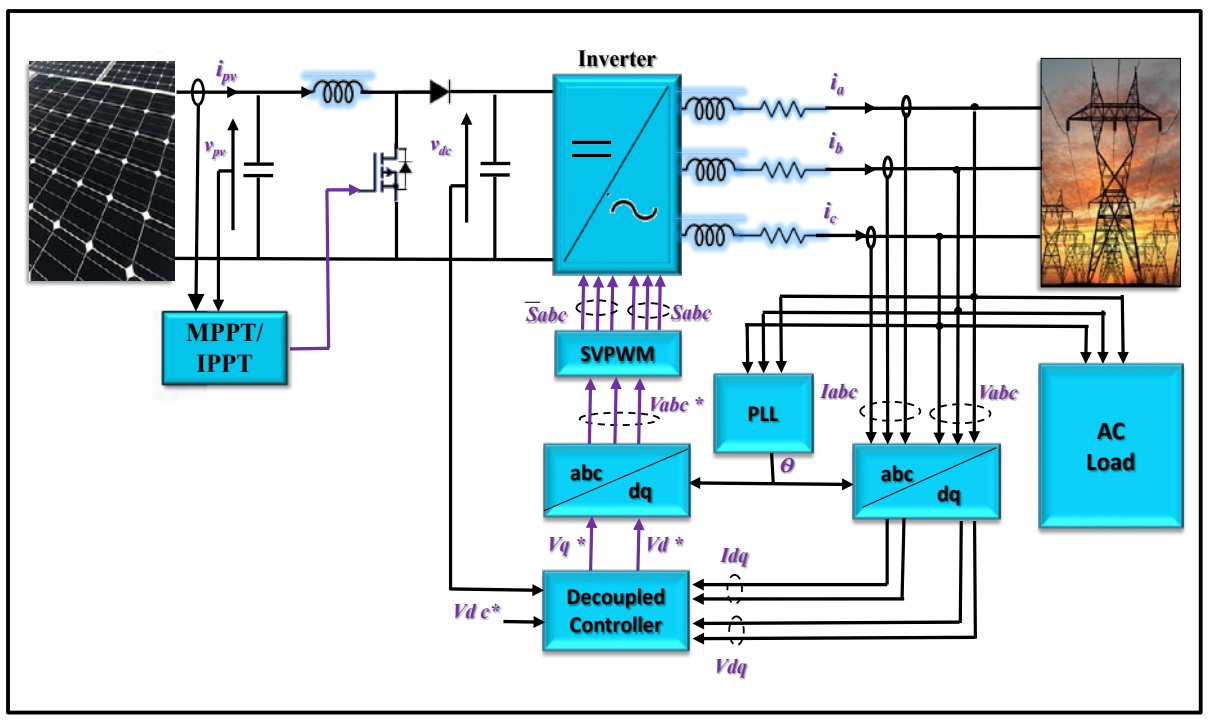

Fig. 2. Setup of the grid-connected PV system by [11], highlighting the signals acquired during the development of the GPVS dataset.

TABLE I. GPVS MPPT DATASET DESCRIPTION

| Filename | Operating State | Data Instances | Complete Cycles |
|---|---|---|---|
| F0M.csv | Normal | 141014 | 705 |
| F1M.csv | Inverter Fault | 139014 | 695 |
| F2M.csv | Feedback Fault | 144015 | 720 |
| F3M.csv | Grid Disturbance | 69967 | 350 |
| F4M.csv | Partial Shading | 144014 | 720 |
| F5M.csv | Open Circuit | 144014 | 720 |
| F6M.csv | Controller fault | 144015 | 720 |
| F7M.csv | Boost Converter fault | 144015 | 720 |

Extracted data from CSV files was cleansed during preprocessing phase by passing noisy signals to the Butterworth low pass filter. The frequency response of this filter is shown in (1) where 'n' represents an order of the filter 'f' represents frequency, and '∈' represents the gain of the bandpass.

$$H(j2\pi f) = \frac{1}{\sqrt{1+\in \left(\frac{2\pi f}{2\pi f_p}\right)^{2n}}} \quad (1)$$

The impact of the Butterworth filter is also shown in Fig. 3, where the noise-corrupted signals and their corresponding filtered signals are illustrated. Moreover, the column representing time in the GPVS dataset is also removed during the preprocessing stage.

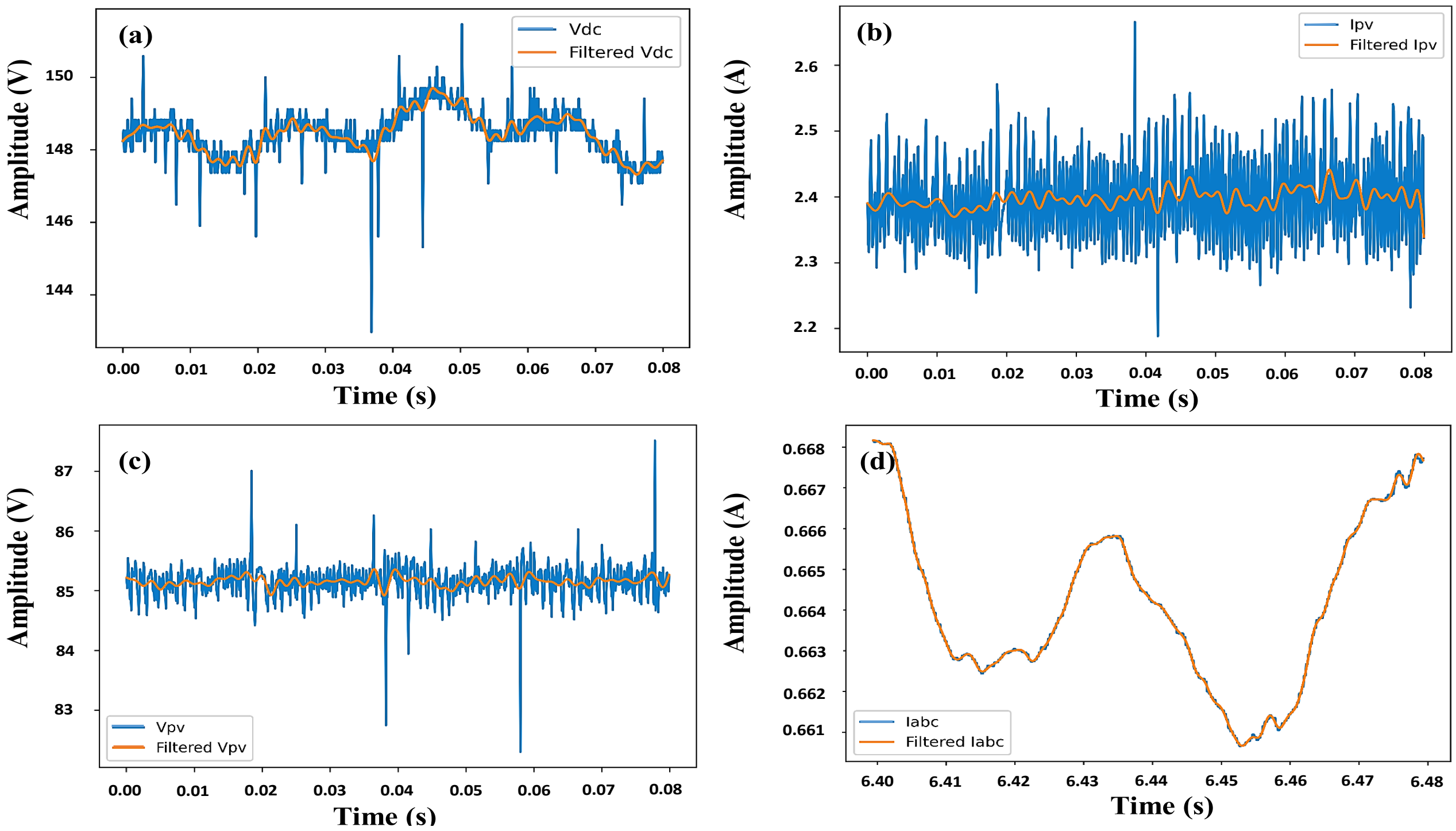

Fig. 3. (a) Actual Vdc and Filtered Vdc Signal (b) Actual Ipv and Filtered Ipv Signal (c) Actual Vpv and Filtered Vpv Signal (d) Actual Iabc and Filtered Iabc Signal

### C. Signatures Development

Consecutive 200 samples of each individual signal in the GPVS dataset represent one complete power cycle and convey more strong information than the individual sample. Therefore, all thirteen signals were divided into batches of 200 consecutive samples. Each individual batch can be utilized for developing the signature of the system's operational state. These signatures were developed using the statistical tools of python. Four simple statistical parameters including mean (m), standard deviation (std), maximum (mx), and minimum (min) amplitude of all batches were calculated. These statistical parameters were utilized in the development of a new dataset, having 52 parameters that were obtained by extracting statistical components from 13 signals of the GPVS dataset. Furthermore, each individual fault sample was labeled at this stage and the resulting dataset of each CSV was concatenated to generate a final dataset having statistical signatures of each signal. For simplicity Pseudocode of the statistical-signatures development process is illustrated in Fig. 4, and mathematical representation of mean and standard deviation is shown in (2) and (3) where 'X' represents individual sample in array of size 'N':

$$Mean = \frac{\sum_{i=1}^{i=N} X_i}{N} \quad (2)$$

$$Standard\ Deviation = \sqrt{\frac{\sum_{i=1}^{N}(X_i - Mean)^2}{N}} \quad (3)$$

**Statistical Signatures Development Process**

```
00:  F0 ←—— Normal Operational State Dataframe
01:  F1 ←—— Fault 1 Dataframe
02:  F2 ←—— Fault 2 Dataframe
03:  F3 ←—— Fault 3 Dataframe
04:  F4 ←—— Fault 4 Dataframe
05:  F5 ←—— Fault 5 Dataframe
06:  F6 ←—— Fault 6 Dataframe
07:  F7 ←—— Fault 7 Dataframe
08:  [F0,F1,F2,F3,F4,F5,F6,F7]←——Slice each dataframe with batches of 200 data instances
09:  for batch in F0:
10:          Class_Label ←—— Set 0 for F0 dataframe
11:          for column in batch:
12:                  calculate mean, standard deviation, maximum and minimum value
13:                  F0_batch_statistics.append( mean, standard deviation, max, min )
14:          end
15:          F0_batch_statistics.append(Class_Label)
16:  end
17:  Calculate batch_statistics for [F1to F7] (similar to Lines 10-17)
18:  F_all_signatures←——concatenate batch_statistics of [F0 to F7]
19:  GPVS_Signature_Dataset←——Create dataframe using F_all_signatures
```

Fig. 4. Pseudocode for extraction of statistical signatures of different operational states of grid-connected PV Systems.

### D. Data Transformation

The GPVS signature dataset is first normalized at the data transformation stage, then it is split into a train-test ratio of 70:30. This training dataset is passed through a random forest classifier initialized with the default parameters of the Scikit learn library [12] for the purpose of features reduction. Out of 52 statistical parameters, the top 30 parameters were selected based on their relative importance and the top priority ones are shown in Fig. 5. Less important features were removed from the train and test split of the dataset before the final training of faults classifier.

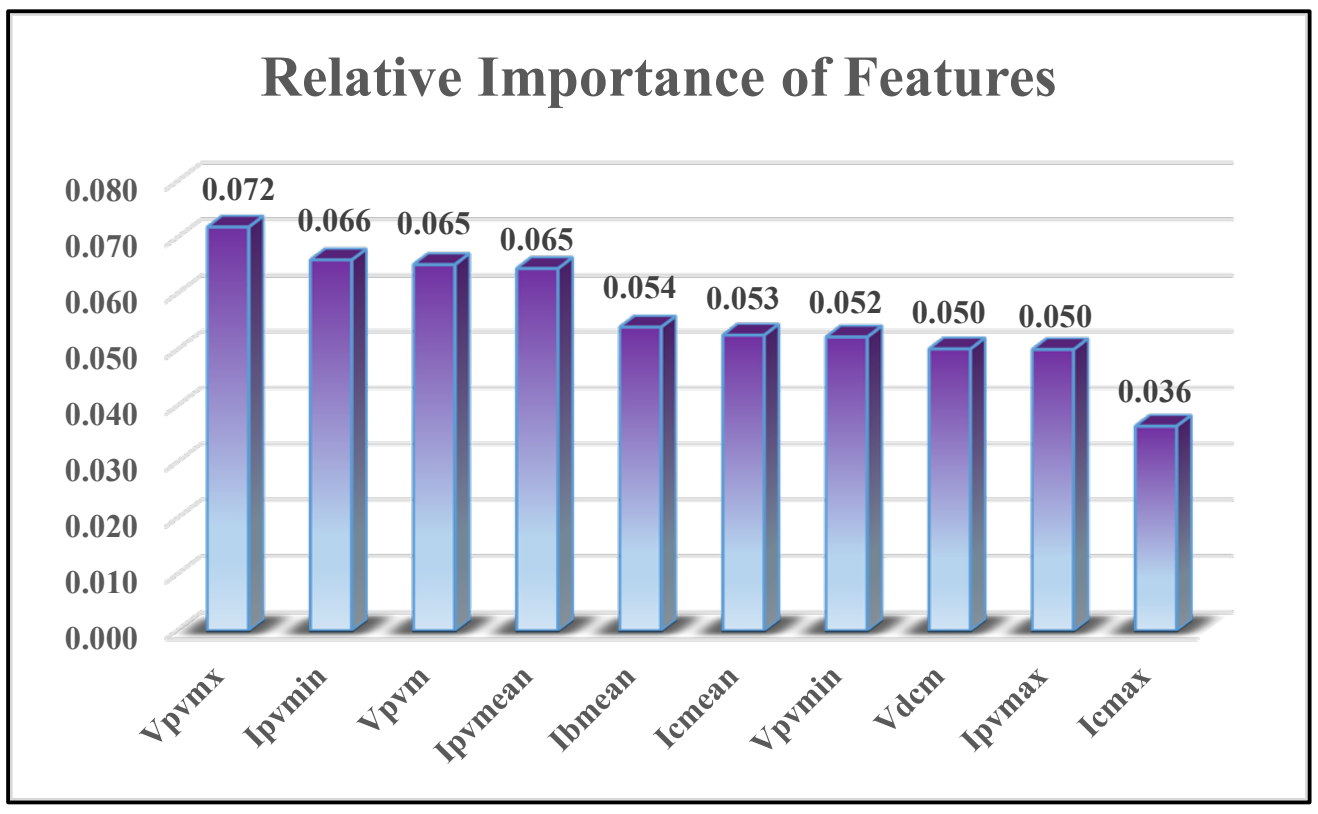

Fig. 5. Relative Feature importance of the grid-connected PV system dataset.

### E. Classifier Training

In this research work, a random forest classifier (RFC) is used for the identification of faults in the PV system because it has shown outclass performance in numerous regression and classification problems [13]. Optimum selection of a number of trees and nodes determines the performance of RFC as they are two essential hyperparameters [14]. In order to select these hyperparameters, the randomized search was utilized with 5-fold cross-validation for preventing the RFC from overfitting problem. The resulting evaluation, as shown in Fig. 6, depicts that 18 trees and log2 are optimum hyperparameters for the classifier.

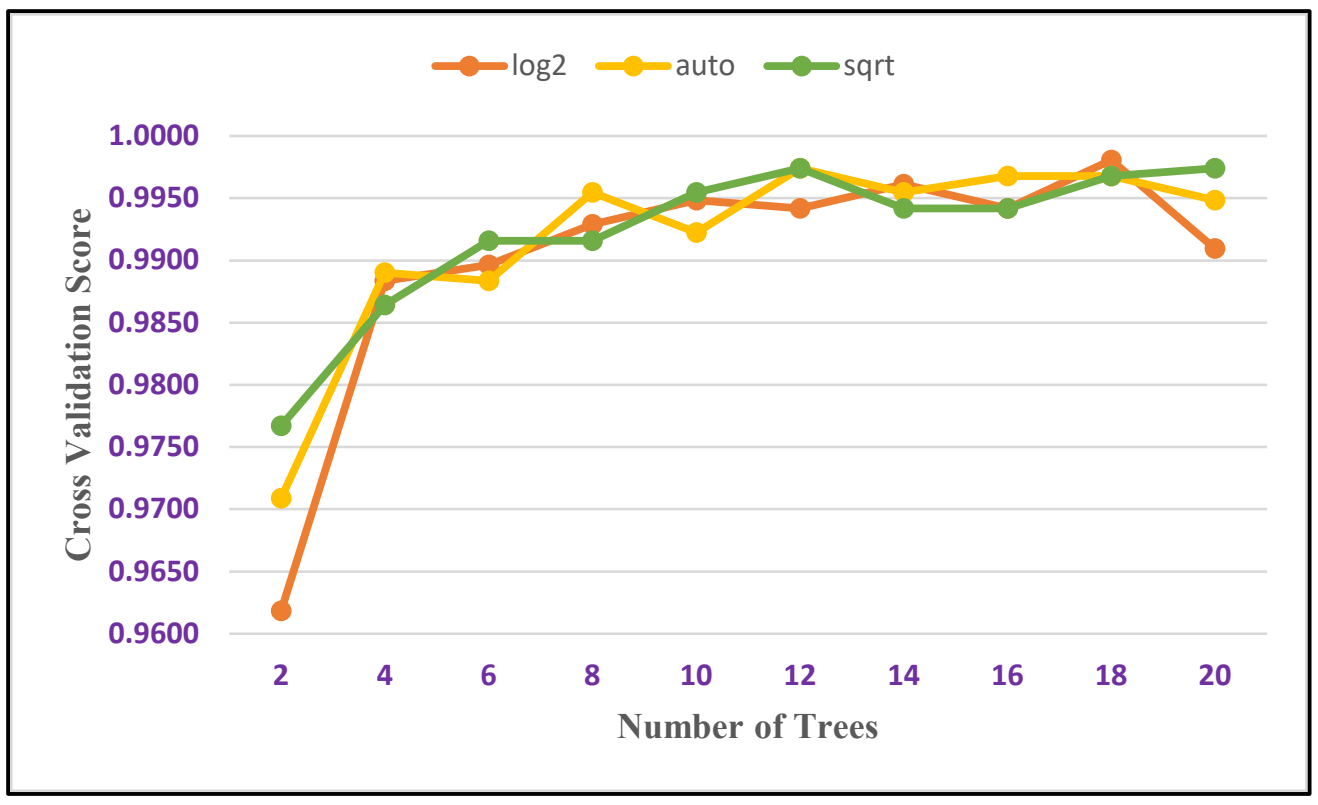

Fig. 6. Results of Random Forest Hyperparametrs randomized search. Thirty RFC models with different numbers of trees and parameters selection criteria were evaluated. The cross-validation score of all possible combinations depicts that 18 trees and log2 are optimum choices for the fault identification system.

### F. SHAP Explainer

RFC models are interpretable than the conventional deep learning models and this interpretation may help to elevate the user trust. This research framework extracts the global explanation of the trained RFC model using SHAP which helps in decoding the behavior of the trained AI model using the game theory approach [10]. Moreover, the tree explainer module of SHAP provides the global and local response of the model with reduced computational complexity [15]. This explanation can also be used for elevating the performance and credibility of ML-based systems [16]. For an illustration of model response and the top decision-making features in the classification of inverter-based faults and sensor-based faults, summary plots of these two classes are shown in Fig. 7.

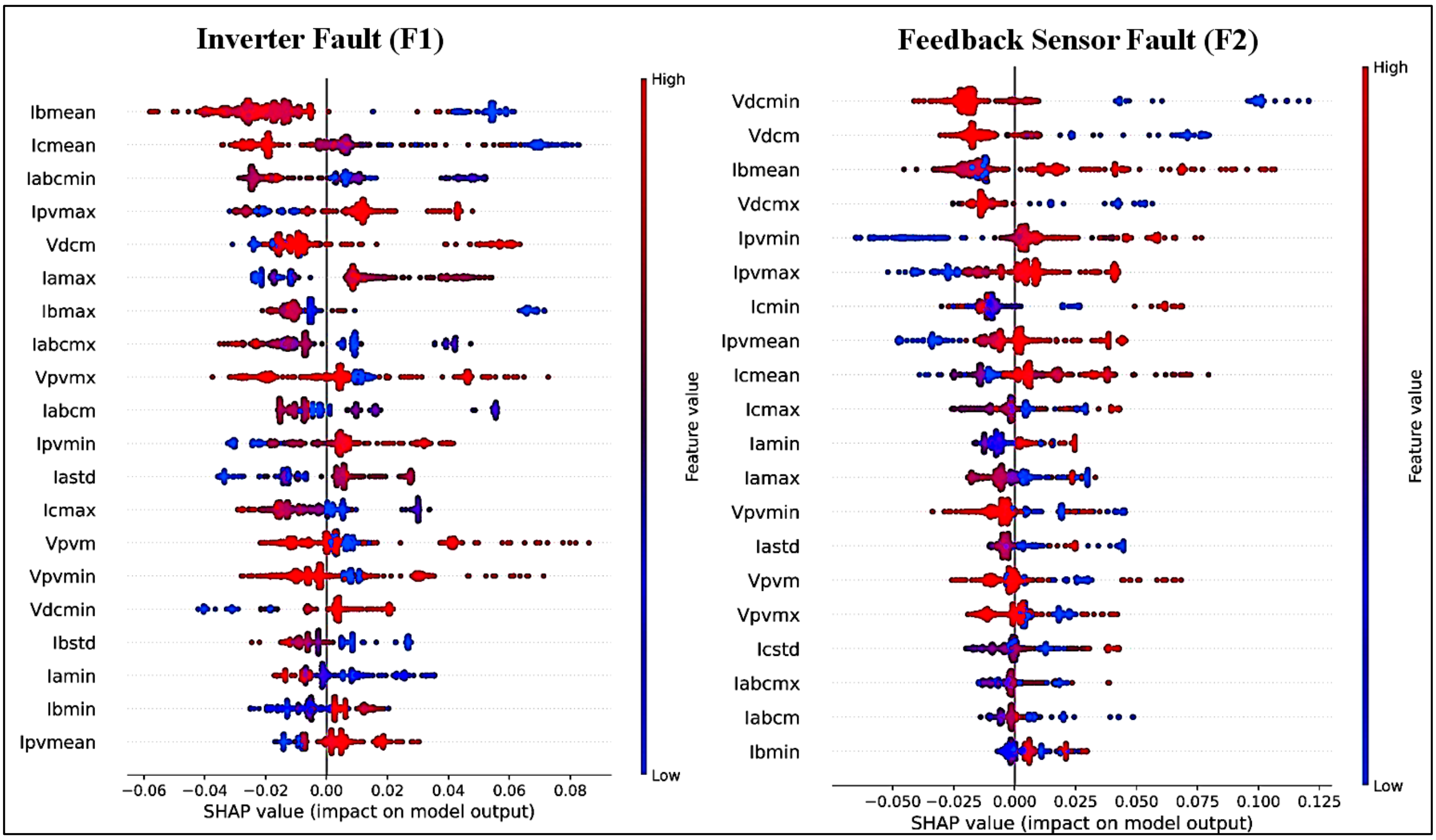

Fig. 7. SHAP Global explanation of Inverter-based and feedback sensor-based faults in the transformed GPVS dataset. The top 20 features extracted from the SHAP summary plots depict their importance and impact in identifying faults in PV systems.

## III. Results and Comparison

The evaluation phase of the proposed Fault Identification System preprocesses the data and then passes this data to the classifier for assessing the operational state of the PV system. Furthermore, this preprocessed data is provided to the SHAP explainer for evaluating the contribution of each feature in the predicted outcome. This additional explainer can be used for enhancing the trust of power operators.

The proposed signature-based fault detection framework is assessed using a test dataset through four evaluation metrics including accuracy, recall, f1-score, and precision. Mathematically these four evaluation metrics are represented as (4), (5), (6), and (7), where TP is True Positive, TN is True Negative, FP is False Positive and FN is False Negative. These evaluation metrics were calculated for the proposed system, Support Vector Machine (SVM), and K-Nearest Neighbors (KNN) classifier of Scikit learn [2]. Their results are shown in Table II and Fig. 8, depicting that the signature-based fault prediction approach outperforms the conventional ML approaches.

$$Accuracy = \frac{TP + TN}{TP + TN + FP + FN} \quad (4)$$

$$Recall = \frac{TP}{TP + FN} \quad (5)$$

$$F1 - Score = \frac{2 * Recall * Precision}{Recall + Precision} \quad (6)$$

$$Precision = \frac{TP}{TP + FP} \quad (7)$$

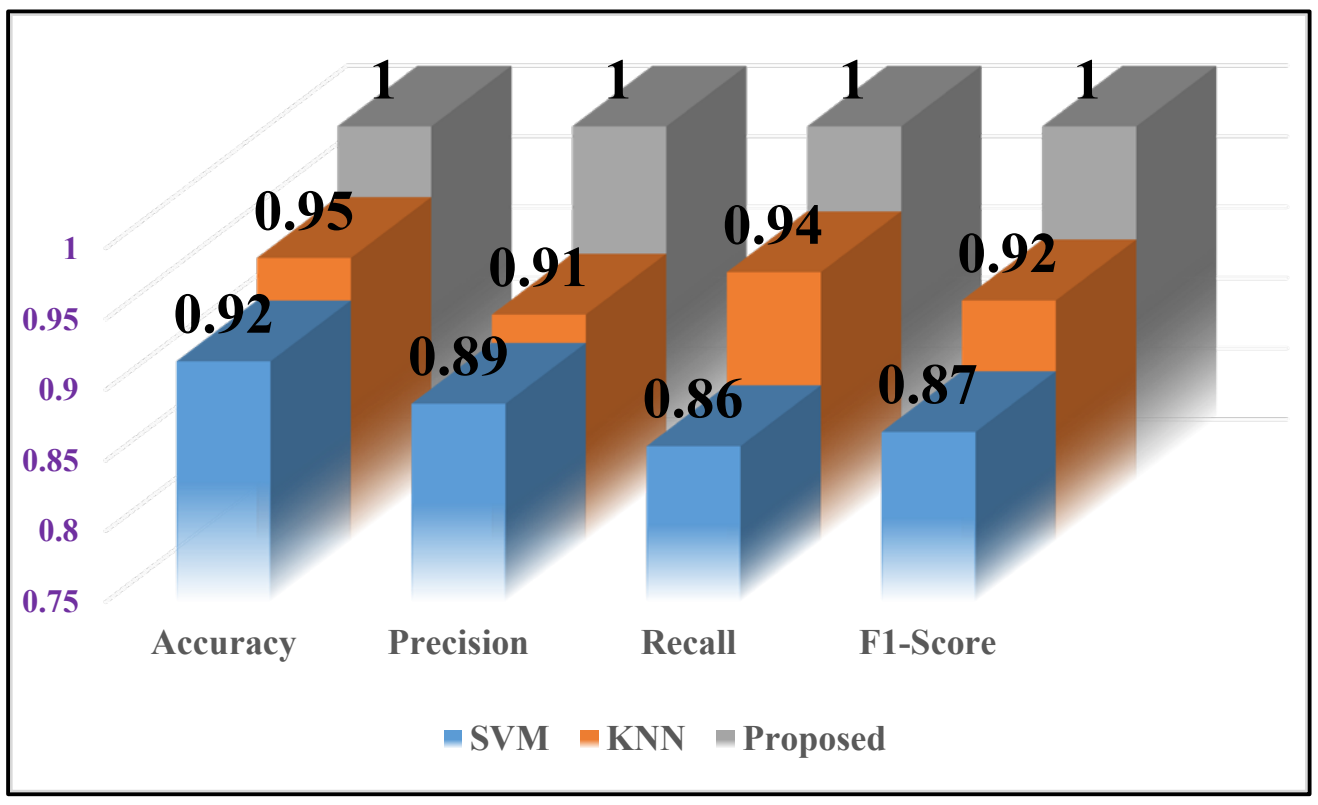

Fig. 8. Macro Average Precision, Recall, F1-Score, Accuracy of SVM, KNN and Proposed Fault Identification System

SHAP based explanation of each individual prediction can also help in the credibility assessment of the predicted outcome. Local explanation of Inverter-based fault (F1M), as shown in Fig. 9, depicts that the prominent features in the decision-making process are present in the feature space of the F1M class. Since the local explanation satisfies the global explanation extracted in the development phase, this features assessment approach can help in the quick validation of all predicted outcomes. Moreover, the presence of top features of any specific class can help the operator to diagnose the corresponding faults in the PV system.

TABLE II. PERFORMANCE EVALUATION ON TEST DATASET

| Faults | KNN | | | | SVM | | | | Proposed IDS | | | |
|---|---|---|---|---|---|---|---|---|---|---|---|---|
| | Accuracy | Precision | Recall | F1-Score | Accuracy | Precision | Recall | F1-Score | Accuracy | Precision | Recall | F1-Score |
| F0M | 0.8333 | 0.69 | 0.83 | 0.76 | 0.1667 | 0.45 | 0.17 | 0.24 | 0.99 | 0.99 | 1 | 1 |
| F1M | 0.9456 | 0.97 | 0.95 | 0.96 | 0.913 | 0.99 | 0.91 | 0.95 | 1 | 1 | 1 | 1 |
| F2M | 0.932 | 0.96 | 0.93 | 0.95 | 0.913 | 1 | 0.91 | 0.95 | 0.998 | 0.99 | 1 | 1 |
| F3M | 1 | 0.75 | 1 | 0.86 | 1 | 1 | 1 | 1 | 1 | 1 | 1 | 1 |
| F4M | 1 | 0.98 | 1 | 0.99 | 1 | 1 | 1 | 1 | 1 | 1 | 1 | 1 |
| F5M | 0.932 | 1 | 0.93 | 0.96 | 0.93 | 0.99 | 0.93 | 0.96 | 1 | 1 | 1 | 1 |
| F6M | 0.954 | 0.97 | 0.95 | 0.96 | 0.99 | 0.81 | 0.99 | 0.89 | 1 | 1 | 1 | 1 |
| F7M | 0.956 | 0.92 | 0.96 | 0.94 | 0.99 | 0.86 | 0.99 | 0.92 | 0.99 | 0.99 | 1 | 1 |

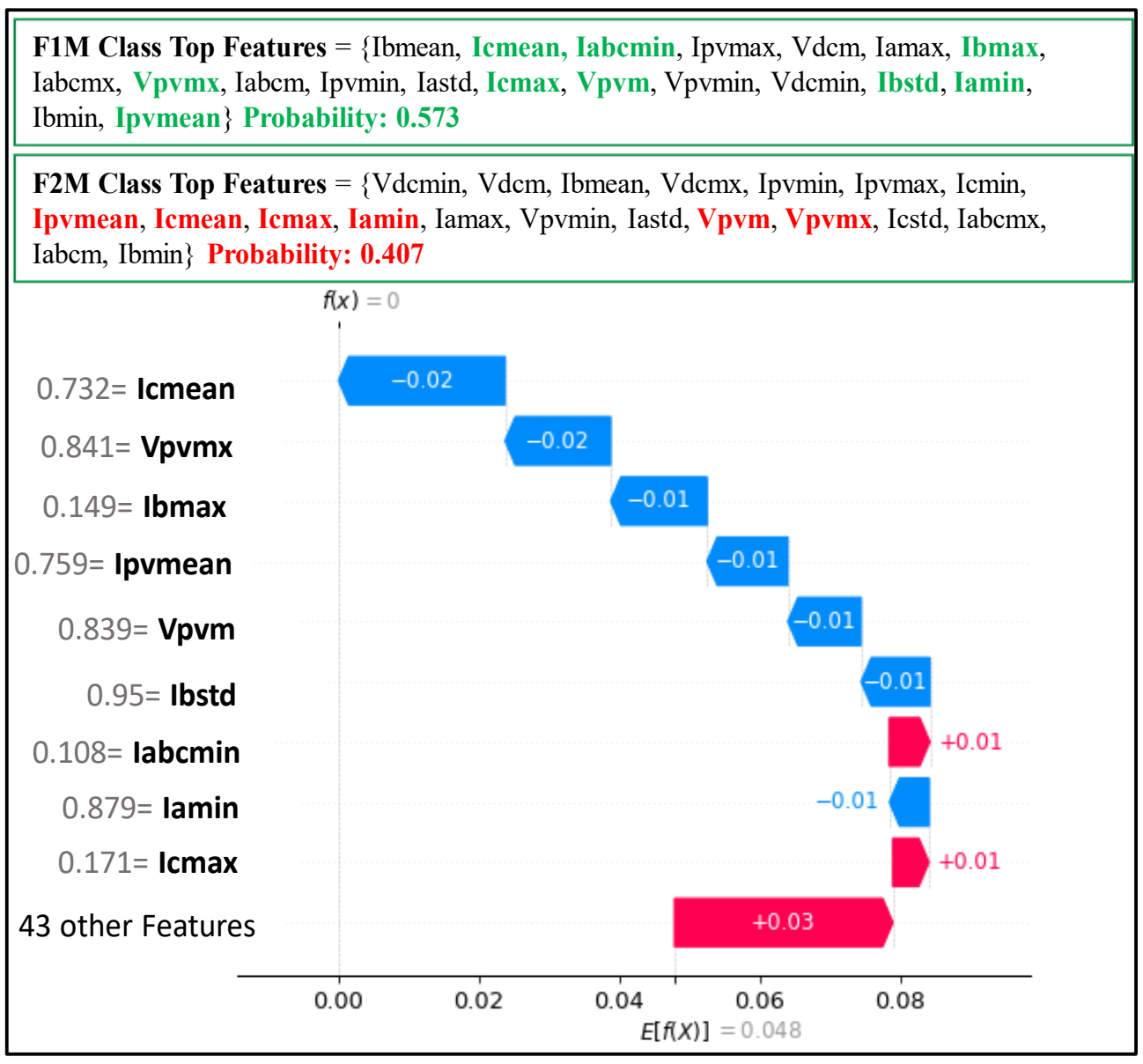

Fig. 9. SHAP waterfall plot for an instance of Inverter-based faults (F1M). Shapley features of two high probability classes are also shown, depicting that the predicted outcome is highly credible because all features in the local explanation are present in the global explanation of the F1M class, whereas only a few of them appear in the global explanation of second high-probability class.

## IV. Conclusion

An increase in power demand and environmental concerns increased the deployment rate of grid-connected PV systems. These systems must operate with high reliability and should be able to detect faults of grid and PV systems to maintain grid stability. This paper addresses this challenge of modern grids by developing a reliable Fault Identification System based on Random Forest Classifier trained on statistical parameters and utilizes the SHAP framework for decoding all predicted outcomes in terms of features contribution. The proposed framework was evaluated on the GPVS dataset using four evaluation metrics and showed 100% accuracy in identifying all types of faults in the GPVS dataset. Furthermore, local explanation of predicted outcome using SHAP adds transparency in the decision-making approach of predicted outcome and can also be utilized for assessing the credibility of the proposed fault identification system.